# Feed Forward and Backward Run in Deep Convolution Neural Network

Pushparaja Murugan

*School of Mechanical and Aerospace Engineering,*

*Nanyang Technological Univeristy, Singapore 639815*

*pushpara001@e.ntu.edu.sg*

## Abstract

Convolution Neural Networks (CNN), known as ConvNets are widely used in many visual imagery application, object classification, speech recognition. After the implementation and demonstration of the deep convolution neural network in Imagenet classification in 2012 by krizhevsky, the architecture of deep Convolution Neural Network is attracted many researchers. This has led to the major development in Deep learning frameworks such as Tensorflow, caffe, keras, theno. Though the implementation of deep learning is quite possible by employing deep learning frameworks, mathematical theory and concepts are harder to understand for new learners and practitioners. This article is intended to provide an overview of ConvNets architecture and to explain the mathematical theory behind it including activation function, loss function, feedforward and backward propagation. In this article, grey scale image is taken as input information image, ReLU and Sigmoid activation function are considered for developing the architecture and cross-entropy loss function is used for computing the difference between predicted value and actual value. The architecture is developed in such a way that it can contain one convolution layer, one pooling layer, and multiple dense layers.

## Nomenclature

$\alpha$      Learning rate

$\hat{y}_i^{L+1}$      Predicated value

| | |
|---|---|
| Ł | Loss or cost function |
| $\sigma$ | Activation function |
| $\sum$ | Summation |
| $a$ | Non-linearly transformed of net input |
| $b$ | Bias- parameter |
| $b^{L+1}$ | Bias matrix of final layer in fully connected layer |
| $b_i^l$ | Bias value of $i^{th}$ neuron at $l^{th}$ layer |
| $C$ | Channel of image |
| $c$ | Depth of convolution kernel |
| $D_1$ | Depth of convolution layer |
| $D_2$ | Depth of pooling layer |
| $D_n$ | Number of pooling layer kernel |
| $Dim_c$ | Dimension of convolution layer |
| $Dim_p$ | Dimension of pooling layer |
| $e$ | Exponential |
| $f'(x)$ | First derivative |
| $f(x)$ | Function |
| $H$ | Width of image |
| $H_1$ | Height of convolution layer |
| $H_2$ | Height of pooling layer |
| $i, j$ | Adjecent neurons in fully connected layer |
| $k$ | Width and height of pooling layer kernel |
| $k^{p,q}$ | Convolution Kernel bank |
| $k_1$ | Width of convolution kernel |
| $k_2$ | Height of convolution kernel |
| $K_D$ | Number of kernel |
| $L$ | Final layers in fully connected layer |
| $l$ | First layers in fully connected layer |
| $L+1$ | Classification layer in fully connected layer |
| $l-1$ | Vectorized pooling layer |
| $n$ | Last neurons in fully connected layer |
| $p$ | Number of convolution kernel |
| $P^{p,q}$ | Pooling Kernel bank |
| $q$ | Number of convolution layer |
| $t$ | Total number of training samples |
| $u, v$ | Pixels of kernel |
| $W$ | Height of image |
| $w$ | Wight- parameter |
| $W^l$ | Wight matrix of first layer in fully connected layer |
| $W^{L+1}$ | Wight matrix of final layer in fully connected layer |
| $W_1$ | Width of convolution layer |
| $W_2$ | Width of pooling layer |
| $w_i^l$ | Wights of $i^{th}$ node at $l^{th}$ layer |
| $x$ | Input signal |
| $y$ | Matrix of actual labled value of training set |
| $y^{L+1}$ | Matrix of predicted value |
| $y_i$ | Actual value from labelled training set |
| $z$ | Linearly transformed net Inputs of fully connected layer |
| $Z_P$ | Value of Zeropadding |
| $Z_S$ | Value of stride |

# 1  Introduction

The study of neural networks, human behavior, and perceptions has started in the early 1950s. Over the decades, different types of neural networks were developed such as Elman, Hopfield and Jordan networks for approximating complex functions and recognizing patterns in the late 1970s. [1] [2] [3]. However, recent development in neural networks profoundly showed incredible results in object classification, pattern recognization, and natural language processing. The advancement in computer vision and the deep Convolution Neural Networks are widely used many application such as cancer cell classification, medical image processing application, star cluster classification, self-driving cars and number plate recognition. CovnNets are bio-inspired artificial neural networks developed on mathematical representation to analyze visual imagery, pattern recognition, and speech recognition. Unlike machine learning, CovnNets can be fed with raw image pixel values rather than feature vectors as input [4]. The basic design principle of CovnNets is developing an architecture and learning algorithm in such way that it reduces the number of the parameter without compromising the computational power of learning algorithm [5]. As the name refers, it consists of the linear mathematical operation of convolution followed by non-linear activators, pooling layers, and deep neural network classifier. The convolution processes act as appropriate feature detectors that demonstrate the ability to deal with a large amount of low-level information. A complete convolution layer has different feature detectors so that multiple features can be extracted from the same image. A single feature detector is smaller in size as compares with the input images is slid over the images for the convolution operation. Hence, all of the units in that feature detector share the same weight and bias. That will help to detect same features in all of the points in the image. That gives the properties of invariance to transformation and shift of the images [6]. Local connections between the pixels are used many times in an architecture. With local respective field, neurons can extract the elementary features such as the orientation of edges and corners and end points. So that higher degree of complex features is detected in hidden layers when its combined in hidden layers. These functions of sparse connectivity between subsequent layers, parameter sharing of weights between the adjacent pixels and equivarient representation enable CNN to use efficiently in image reorganization and image classification problems [7] [8].

## 2 Architecture

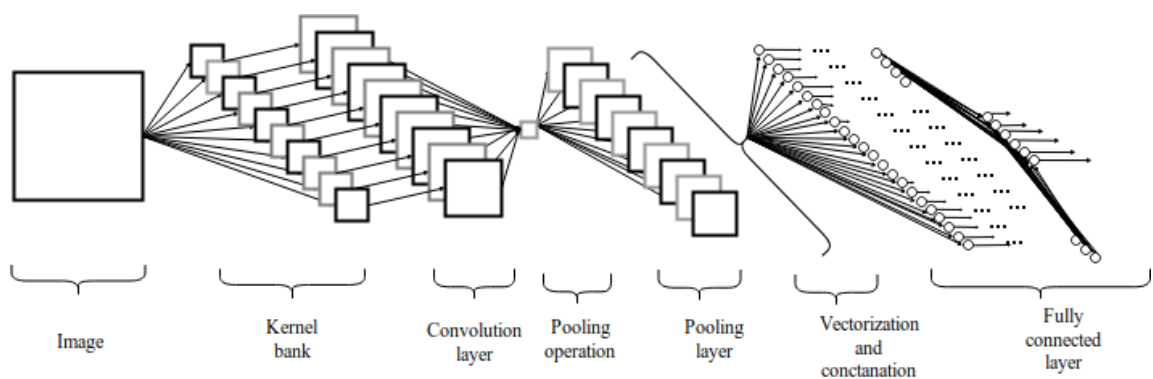

Figure 2.1: Architecture Convolution Neural Network

### 2.1 Convolution layers

Convolution layers are set of parallel feature maps, formed by sliding different kernel (feature detector) over an input image and projecting the element-wise dot as the feature maps [9]. This sliding process is known as stride $Z_s$. This kernel bank is smaller in size as compares with the input image and are overlapped on the input image which prompts the parameters such as weight and bias sharing between the adjacent pixel of the image as well as control the dimensions of feature maps. Using the small size of kernels, however often result in imperfect overlays and limit the power of the learning algorithm. Hence, Zero padding $Z_p$ process usually implemented to control the size of the input image. Zero padding will control the feature maps and kernels dimensions independently by adding zero to input symmetrically [10]. During the training of algorithm, set of kernel filters, known as filter bank with the dimension of $(k_1, k_2, c)$, slide over the fixed size $(H, W, C)$ input image. The stride and zero padding are the critical measures to control the dimension of the convolution layers. As a result feature maps are produced which are stacked together to form the convolution layers. The dimension of the convolution layer can be computed by following Eqn. 2.1.

$$Dim_c(H_1, W_1, D_1) = (H + 2Z_P - k_1)/Z_S + 1), (W + 2Z_P - k_2)/Z_S + 1), K_D \quad \text{(Eq. 2.1)}$$

## 2.2 Activation functions

Activation function defines the output of a neuron based on given a set of inputs. Weighted sum of linear net input value is passed through an activation function for non-linear transformation. A typical activation function is based on conditional probablity which will return the value one or zero as a output $op$ $\{P(op = 1|ip)$ or $P(op = 0|ip)\}$. When the net input information $ip$ cross the threshold value, the activation function returns to value one and it passes the information to the next layers. If the net input $ip$ value is below the threshold value, it returns to value zero and will not pass the information. Based on this segregation of relevent and irrelevent information, the activation function decides whether the neuron should activate or not. Higher the net input value greater the activation. Different types of activation functions are developed and used for different application. Some of the commonly used activation function are given in the Table 1.

## 2.3 Pooling layers

Pooling layer refers to downsampling layer which combines the output of the neuron cluster at one layer to single neuron in the next layer. Pooling operations carried out after the non-linear activation where the pooling layers help to reduce the number of data points and to avoid overfitting. It also act as a smoothing process from which unwanted noise can be eliminated. Most commonly Max pooling operation is used. Addition to that average pooling and $L_2$ norm pooling operation are also used in some cases.

When $D_n$ number of kernel windows and the stride value of $Z_S$ is employed to develop pooling layers, the dimension of the pooling layer can be computed by,

$$Dim_p(H_2, W_2, D_2) = (H_1 - k)/Z_S + 1), (W_1 - k)/Z_S + 1), D_n \quad \text{(Eq. 2.2)}$$

## 2.4 Fully connected dense layers

After the pooling layers, pixels of pooling layers is stretched to single column vector. These vectorized and concatnated data points are fed into dense layers ,known as fully connected layers for the classification. The function of fully connected dense layers is similar to Deep Neural Neworks. The architecture of CovnNets is given in Figure 2.1. This type of constraint architecture will proficiently surpass the classical machine learning algorithms in image classification problems [11] [12].

## 2.5 Loss or cost function

Loss function maps an event of one or more variable onto a real number associated with some cost. Loss function is used to measure the performance of the model and inconsistency between actual $y_i$ and predicted value $\hat{y}_i^{L+1}$. Performance of model increses with the decrease value of loss function.

| Name | Functions | Derivatives | Figure |
|------|-----------|-------------|--------|
| **Sigmoid** | $\sigma(x) = \frac{1}{1+e^{-x}}$ | $f'(x) = f(x)(1-f(x))^2$ | 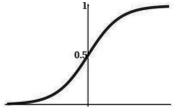 |
| **tanh** | $\sigma(x) = \frac{e^x - e^{-x}}{e^z + e^{-z}}$ | $f'(x) = 1 - f(x)^2$ | 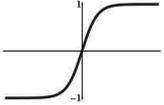 |
| **ReLU** | $f(x) = \begin{cases} 0 & \text{if } x < 0 \\ x & \text{if } x \geq 0. \end{cases}$ | $f'(x) = \begin{cases} 0 & \text{if } x < 0 \\ 1 & \text{if } x \geq 0. \end{cases}$ | 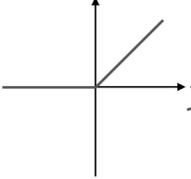 |
| **Leaky ReLU** | $f(x) = \begin{cases} 0.01x & \text{if } x < 0 \\ x & \text{if } x \geq 0. \end{cases}$ | $f'(x) = \begin{cases} 0.01 & \text{if } x < 0 \\ 1 & \text{if } x \geq 0. \end{cases}$ | 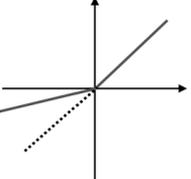 |
| **Softmax** | $f(x) = \frac{e^x}{\sum_1^j e^x}$ | $f'(x) = \frac{e^x}{\sum_1^j e^x} - \frac{(e^x)^2}{(\sum_1^j e^x)^2}$ | |

Table 1: Non-linear activation function

If the output vector of all possible output is $y_i = \{0, 1\}$ and an event $x$ with set of input vector variable $x = (x_i, x_2 \ldots x_t)$, then the mapping of $x$ to $y_i$ is given by,

$$\text{L}(\hat{y}_i^{L+1}, y_i) = \frac{1}{t} \sum_{i=1}^{i=t} (y_i, (\sigma(x), w, b)) \qquad \text{(Eq. 2.3)}$$

where $\text{L}(\hat{y}_i^{L+1}, y_i)$ is loss function
Many types of loss functions are developed for various applications and some are given below.

### 2.5.1 Mean Squared Error

Mean Squared Error or known as quadratic loss function, is mostly used in linear regression models to measure the performance. If $\hat{y}_i^{L+1}$ is the computed output value of t training sample and $y_i$ is the corresponding labeled value, then the Mean Squared Error(MSE) is given by,

$$\text{L}(\hat{y}_i^{L+1}, y_i) = \frac{1}{t} \sum_{i=1}^{i=t} (y_i - \hat{y}_i^{L+1})^2 \qquad \text{(Eq. 2.4)}$$

Downside of the MSE is, tends to suffer from slow learning speed (slow convergence) when it incorprated with Sigmoid activation function.

### 2.5.2 Mean Squared Logarithmic Error

Mean Squared Logarithmic Error(MSLE) is also used to measure performance of the model.

$$Ł(\hat{y}_i^{L+1}, y_i) = \frac{1}{t}\sum_{i=1}^{i=t}\left(\log(y_i+1) - \log(\hat{y}_i^{L+1}-1)\right)^2 \qquad \text{(Eq. 2.5)}$$

### 2.5.3 $L_2$ Loss function

$L_2$ loss function is square root of $L_2$ norm of the difference between actual labeled value and computed value from the net input and is given by,

$$Ł(\hat{y}_i^{L+1}, y_i) = \sum_{i=1}^{i=t}(y_i - \hat{y}_i^{L+1})^2 \qquad \text{(Eq. 2.6)}$$

### 2.5.4 $L_1$ Loss function

L1 loss function is sum of absolute errors of the difference between actual labeled value and computed value from the net input and is expressed as,

$$Ł(\hat{y}_i^{L+1}, y_i) = \sum_{i=1}^{i=t}|y_i - \hat{y}_i^{L+1}| \qquad \text{(Eq. 2.7)}$$

### 2.5.5 Mean Absolute Error

Mean Absolute Error is used to measure the proximity of the predictions and actual values, which is expressed by,

$$Ł(\hat{y}_i^{L+1}, y_i) = \frac{1}{t}\sum_{i=1}^{i=t}|y_i - \hat{y}_i^{L+1}| \qquad \text{(Eq. 2.8)}$$

### 2.5.6 Mean Absolute Percentage Error

Mean Absolute Percentage Error is given by,

$$Ł(\hat{y}_i^{L+1}, y_i) = \frac{1}{t}\sum_{i=1}^{i=t}|(\frac{y_i - \hat{y}_i^{L+1}}{y_i})| \times 100 \qquad \text{(Eq. 2.9)}$$

Major downside of MAPE is, inablity to perform when there are zero values.

### 2.5.7 Cross Entrophy

The most commonly used loss function is Cross Entropy loss function and is expained below. If the probablity of output $y_i$ is in the traning set label $y_i^{\tilde{L}+1}$ is, $P(y_i|a^{l-1}) = i_t^{\hat{L}+1} = 1$ and the the probablity of output $y_i$ is not in the traning set label $y_i^{\hat{L}+1}$ is, $P(y_i|z^{l-1}) = y_i^{\hat{L}+1} = 0$ [13]. The expected label is $y$, than Hence,

$$P(y_i|z^{l-1}) = \hat{y}_i^{L+1} y_i (1 - \hat{y}_i^{L+1})^{(1-y_i)} \tag{Eq. 2.10}$$

$$\log P(y_i|z^{l-1}) = log((\hat{y}_i^{L+1})^{(y_t)}(1 - \hat{y}_i^{L+1})^{(1-y_i)}) \tag{Eq. 2.11}$$

$$= (y_i)\log(\hat{y}_i^{L+1}) + (1 - y_i)\log(1 - \hat{y}_t^{L+1}) \tag{Eq. 2.12}$$

To minimize the cost function,

$$\log P(y_i|z^{l-1}) = -log((\hat{y}_i^{L+1})^{(y_i)}(1 - \hat{y}_i^{L+1})^{(1-y_i)}) \tag{Eq. 2.13}$$

In case of $i$ training samples, the cost function is,

$$Ł(\hat{y}_i^{L+1}, y_i) = -\frac{1}{t}\sum_{1}^{i=t}((y_i)\log(\hat{y}_i^{L+1}) + (1 - y_i)\log(1 - \hat{y}_i^{L+1})) \tag{Eq. 2.14}$$

$$\tag{Eq. 2.15}$$

## 3 Learning of CovnNets

### 3.1 Feed - Forward run

Feed forward run or propogation can be explained as mutiplying the input value by randomly initiated weights and adding randomly initiated bias values of each connection of every neurons followed by summation of all the products of all the neurons. Then passing the net input value through non-linear activation functions.

In a discrete color space, image and kernel can be represented as a 3D tensor with the dimension of $(H, W, C)$ and $(k_1, k_2, c)$ where $m, n, c$ are represent the $m^{th}, n^{th}$ pixel in $c^{th}$ channel. First two indices are indicate the spatial co-ordinates and last index is indicate the color channel.

If a kernel is slided over the color image, the multidimensional tensor convolution operation can be expressed as,

$$(I \bigotimes K)_{ij} = \sum_{m=1}^{m}\sum_{n=1}^{n}\sum_{c=1}^{C} K_{m,n,c} I_{i+m,j+n,c} \tag{Eq. 3.1}$$

Convolution process is indicated by $\bigotimes$ sympol.
For grey scale image, convolution process can be expressed as,

$$(I \bigotimes K)_{ij} = \sum_{m=1}^{m}\sum_{n=1}^{n} K_{m,n} I_{i+m,j+n} \tag{Eq. 3.2}$$

A kernel bank $k_{u,v}^{p,q}$ is slided over the image $I_{m,n}$ with stride value of 1 and zero padding value of 0. The feature maps of the convolution layer $C_{m,n}^{p,q}$ can be computed by,

$$C_{m,n}^{p,q} = \sum_{m=1}^{m}\sum_{n=1}^{n} I_{(m-u,n-v)}.K_{u,v}^{p,q} + b^{p,q} \tag{Eq. 3.3}$$

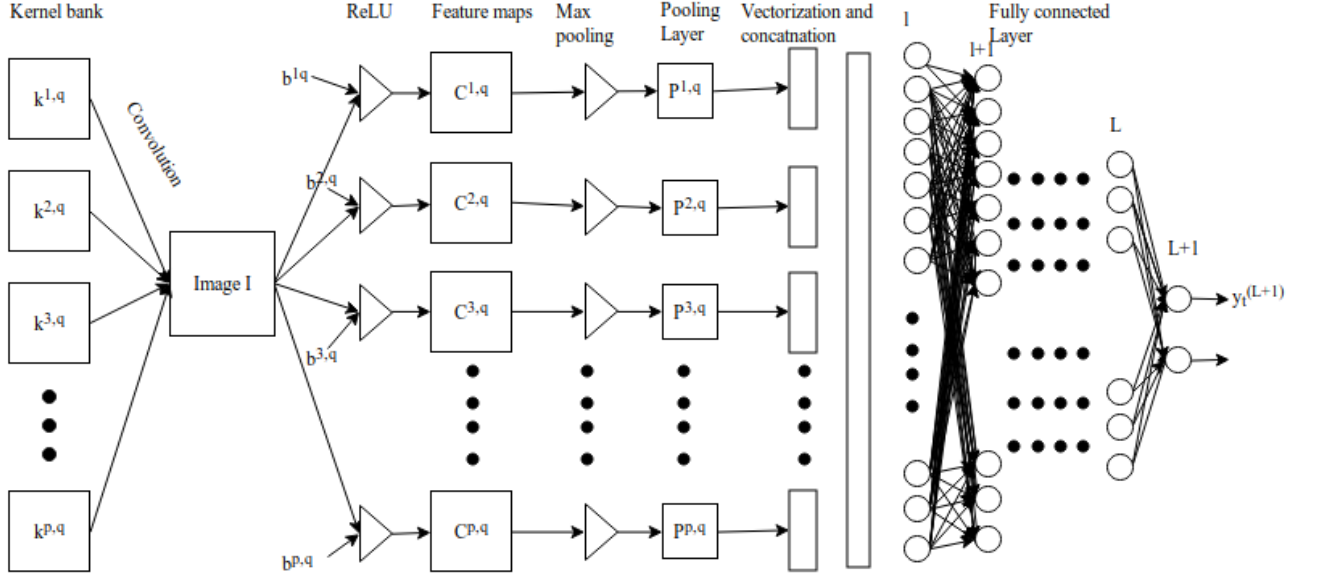

Figure 3.1: Convolution Neural Network

These feature maps are passed through a non-linear acitivation function $\sigma$,

$$C^{p,q}_{m,n} = \sigma(\sum_{m=1}^{m}\sum_{n=1}^{n} I_{(m-u,n-v)} \cdot K^{p,q}_{u,v} + b^{p,q}) \qquad \text{(Eq. 3.4)}$$

where $\sigma$ is a ReLU activation fucntion.
Pooling layer $P^{p,q}_{m,n}$ is developed by taking out the maximum valued pixels $m,n$ in the convolution layers. The pooling layer can be calculated by,

$$P^{p,q}_{m,n} = max(C^{p,q}_{m,n}) \qquad \text{(Eq. 3.5)}$$

The pooling layer $P^{p,q}$ is concatenated to form a long vector with the length of $p \times q$ and is fed into fully connected dense layers for the classification, then the vecotoized data points $a^{l-1}_i$ in $l-1$ layer is given by,

$$a^{l-1}_i = f(P^{p,q}) \qquad \text{(Eq. 3.6)}$$

This long vector is fed into a fully connected dense layers from $l$ layer to $L+1$. If the fully connected dense layers is developed with $L$ number of layers and $n$ number of neurons, then $l$ is the first layer, $L$ is the last layer and $(L+1)$ is the classification layer as shown in the figure 3.2, the forward run between the layers are given by,

$$z^l_1 = w^l_{11}a^{l-1}_1 + w^l_{12}a^{l-1}_2 + \cdots + w^{l-1}_{1j}a^{l-1}_1 + \cdots + b^l_j \qquad \text{(Eq. 3.7)}$$

$$z^l_2 = w^l_{21}a^{l-1}_1 + w^l_{22}a^{l-1}_2 + \cdots + w^{l-1}_{2j}a^{l-1}_1 + \cdots + b^l_j \qquad \text{(Eq. 3.8)}$$

$$z^l_i = w^l_{i1}a^{l-1}_i + w^l_{ij}a^{l-1}_i + \cdots + w^{l-1}_{2j}a^{l-1}_i + \cdots + b^l_j \qquad \text{(Eq. 3.9)}$$

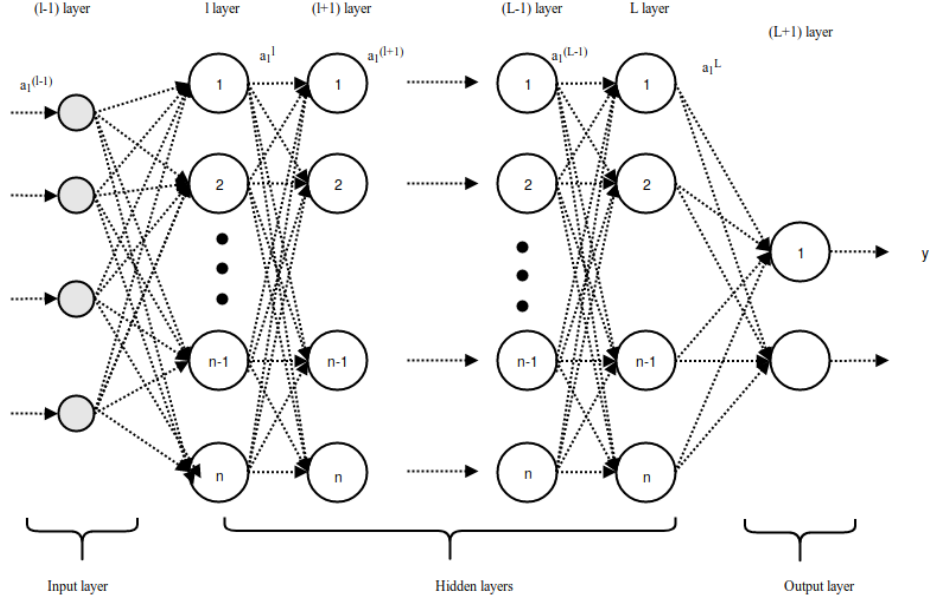

Figure 3.2: Forward run in fully connected layer

$$\begin{bmatrix} z_1^l \\ \vdots \\ z_i^l \\ \vdots \end{bmatrix} = \begin{bmatrix} w_{11}^l & w_{12}^l & w_{13}^l & \cdots & w_{1n}^l \\ \vdots & \vdots & \vdots & \vdots & \vdots \\ w_{i1}^l & w_{i2}^l & w_{i3}^l & \cdots & w_{in}^l \\ \vdots & \vdots & \vdots & \ddots & \vdots \end{bmatrix} \begin{bmatrix} a_1^{l-1} \\ \vdots \\ a_i^{l-1} \\ \vdots \end{bmatrix} + \begin{bmatrix} b_1^l \\ \vdots \\ b_i^l \\ \vdots \end{bmatrix} \qquad \text{(Eq. 3.10)}$$

Consider a single neuron $(j)$ in a fully connected layer at layer $l$ as given in the Fig.3.3. The input values $a_i^{l-1}$ are multiplied and added by weights $w_{ij}$ and bias values $b_j^l$ respectively. Then the final net input value $z_i^l$ are passed through a non-linear activation function $\sigma$. Then the corresponding output value $a_j^l$ is computed by,

$$z_j^l = w_{1j}^l a_1^{l-1} + w_{2j}^l a_2^{l-1} + \cdots + w_{ij}^{l-1} a_i^{l-1} + \cdots + b_j^l \qquad \text{(Eq. 3.11)}$$

Where $z_i^l$ is the input of the activation function for the neuron $j$ at layer $l$,

$$z_j^l = \sum_{i=1}^{n} w_{ij}^l a_j^{l-1} + b_i^l \qquad \text{(Eq. 3.12)}$$

$$a_j^l = \sigma(\sum_{i=1}^{n} w_{ij}^l a_j^{l-1} + b_i^l) \qquad \text{(Eq. 3.13)}$$

$$\text{(Eq. 3.14)}$$

Hence, the output of $l^{th}$ layer is,

$$a^l = \sigma((W^l)^T a^{l-1} + b^l) \qquad \text{(Eq. 3.15)}$$

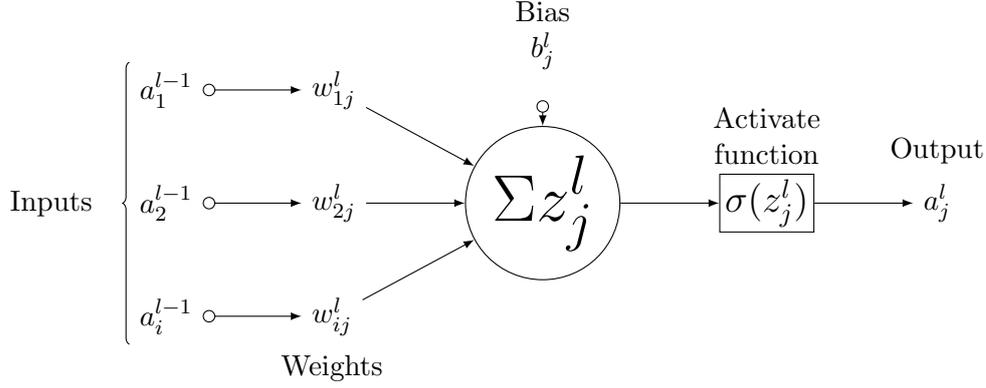

Figure 3.3: Forward run in a neuron $j$ at $l^th$ layer

$$a^l = \sigma(z^l) \tag{Eq. 3.16}$$

where $a^l$ is,

$$a^l = \begin{bmatrix} a_1^l \\ \vdots \\ a_i^l \\ \vdots \end{bmatrix} = \begin{bmatrix} \sigma(z_1^l) \\ \vdots \\ \sigma(z_i^l) \\ \vdots \end{bmatrix} \tag{Eq. 3.17}$$

$W^l$ is,

$$W^l = \begin{bmatrix} w_{1j}^l \\ \vdots \\ w_{ij}^l \\ \vdots \end{bmatrix} \tag{Eq. 3.18}$$

In this same manner, the output value of last leyer $L$ is given by,

$$a^L = \sigma((W^L)^T a^{L-1} + b^L) \tag{Eq. 3.19}$$

where,

$$a^L = \sigma(z^L) \tag{Eq. 3.20}$$

$$a^L = \begin{bmatrix} a_1^L \\ \vdots \\ a_i^L \\ \vdots \end{bmatrix} = \begin{bmatrix} \sigma(z_1^L) \\ \vdots \\ \sigma(z_i^L) \\ \vdots \end{bmatrix} \tag{Eq. 3.21}$$

Expanding this to classification layers, final output predicted value $\hat{y}_i^{L+1}$ of a neuron unit ($i$) at $L + 1$ layer can be expressed as,

$$\hat{y}_i^{L+1} = \sigma(W^L \ldots \ldots \sigma(W^2(\sigma(W^1 a^1 + b^1) + b^2 \ldots \cdots + b^L)) \tag{Eq. 3.22}$$

If the predicted value is $\hat{y}_i^{L+1}$ and the actual labeled value is $y_i$, than the performance of the model can be computed by the following loss function equation,
From the Eqn.2.14, cross-entropy loss function is,

$$L(\hat{y}_i^{L+1}, y_i) = -\frac{1}{t}\sum_{1}^{i=t}((y_i)\log(\hat{y}_i^{L+1}) + (1-y_i)\log(1-\hat{y}_i^{L+1})) \qquad \text{(Eq. 3.23)}$$

(Eq. 3.24)

## 3.2 Backward run

Backward run, also known as backward propogation is referred to backward propogation of errors which use gradient descent to compute the gradient of the loss function with respect to the parameters such as weight and bias and is shown in the Fig 3.4. During the backward propogation, gradient of loss function of final layers with respect to the parameters is computed first where the gradient of first layer is computed last. Also, the partial derivative of one layers is reused in computation of partial derivative of another layers by chain rule which will lead to efficient computation of gradient at each layers. This will be used to minimize the loss function. Performance of model increases as the loss function value decreses [14] [15] [16].

In the back propogation, the paramters such as $W^{L+1}, b^{L+1}, W^l, b^l,, k^{p,q}$ and $b^{p,q}$ are needed to be update in order to minimize the cost function.

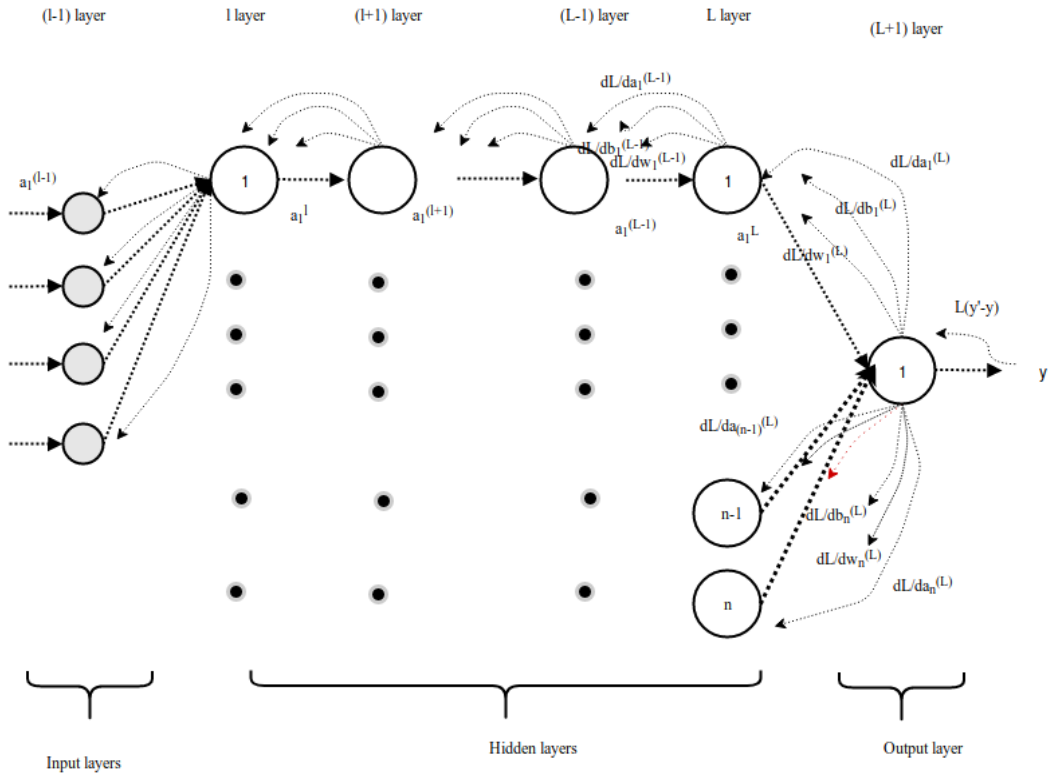

Figure 3.4: Back propogation in fully connected layer

Partial derivative of loss function of $i^{th}$ neuron at classification layer $L+1$ with respect to predicted values $\hat{y}_i^{L+1}$ is,

$$\frac{\partial \text{L}(\hat{y}_i^{L+1}, y_i)}{\partial y_i^{L+1}} = \frac{1}{t}\sum_1^{i=t} \frac{\partial(-1((y_t \log(\hat{y}_i^{L+1}) + (1-y_i)\log((l-\hat{y}_i^{L+1})))}{\partial \hat{y}_i^{L+1}} \qquad \text{(Eq. 3.25)}$$

$$\frac{\partial \text{L}(\hat{y}_i^{L+1}, y_i)}{\partial \hat{y}_i^{L+1}} = \frac{1}{t}\sum_1^{i=t} \frac{-y_i}{\hat{y}_i^{L+1}} + \frac{1-y_i}{1-\hat{y}_i^{L+1}} \qquad \text{(Eq. 3.26)}$$

In case of multiclass categorical classification problem, the lost function of classification layer $L+1$ is,

$$\begin{bmatrix} \frac{\partial \text{L}(\hat{y}_1^{L+1}, y_1)}{\partial \hat{y}_1^{L+1}} \\ \frac{\partial \text{L}(\hat{y}_2^{L+1}, y_2)}{\partial \hat{y}_2^{L+1}} \\ \vdots \\ \frac{\partial \text{L}(y_i^{L+1}, y_i)}{\partial \hat{y}_i^{L+1}} \\ \vdots \end{bmatrix} = \begin{bmatrix} \frac{1}{i}\sum_1^t \frac{-y_1}{\hat{y}_1^{L+1}} + \frac{1-y_1}{1-\hat{y}_1^{L+1}} \\ \frac{1}{i}\sum_1^t \frac{-y_2}{\hat{y}_2^{L+1}} + \frac{1-y_2}{1-\hat{y}_2^{L+1}} \\ \vdots \\ \frac{1}{i}\sum_1^t \frac{-y_i}{\hat{y}_i^{L+1}} + \frac{1-y_i}{1-\hat{y}_i^{L+1}} \\ \vdots \end{bmatrix} \qquad \text{(Eq. 3.27)}$$

Partial derivate of cost function with respect to weight $w_{i,i-1}^{L+1}$ of $i^{th}$ neuron in final layer $L$, For convinent purpose, the notation of the weight of $L^{th}$ layer is denoted as $w_{i,i-1}^L$.

$$\frac{\partial \text{L}(\hat{y}_i^{L+1}, y_i)}{\partial w_{i,i-1}^{L+1}} = \frac{1}{t}\sum_1^{i=t} \frac{\partial \text{L}(\hat{y}_i^{L+1}, y_i)}{\partial \hat{y}_i^{L+1}} \frac{\partial \hat{y}_i^{L+1}}{\partial w_{i,i-1}^{L+1}} \qquad \text{(Eq. 3.28)}$$

$$= \frac{1}{t}\sum_1^t (\frac{-y_i}{\hat{y}_i^{L+1}} + \frac{1-y_i}{1-\hat{y}_i^{L+1}})(\frac{\partial \hat{y}_i^{L+1}}{\partial w_{i,i-1}^{L+1}}) \qquad \text{(Eq. 3.29)}$$

$$= \frac{1}{t}\sum_1^t (\frac{-y_i}{\hat{y}_i^{L+1}} + \frac{1-y_i}{1-\hat{y}_i^{L+1}})(\frac{\partial a_t^{L+1}}{\partial w_{i,i-1}^{L+1}}) \qquad \text{(Eq. 3.30)}$$

$$= \frac{1}{t}\sum_1^t (\frac{-y_i}{\hat{y}_i^{L+1}} + \frac{1-y_i}{1-\hat{y}_i^{L+1}})(\frac{\partial \sigma(z_t^{L+1})}{\partial w_{i,i-1}^{L+1}}) \qquad \text{(Eq. 3.31)}$$

$$= \frac{1}{t}\sum_1^t (\frac{-y_i}{\hat{y}_i^{L+1}} + \frac{1-y_i}{1-\hat{y}_i^{L+1}})\sigma'(z_i^{L+1}) \qquad \text{(Eq. 3.32)}$$

$$= \frac{1}{t}\sum_1^t (\frac{-y_i}{\hat{y}_i^{L+1}} + \frac{1-y_i}{1-\hat{y}_i^{L+1}})\sigma'(\sum_{i=1}^i w_{i,i-1}a^{L-1} + b^L) \qquad \text{(Eq. 3.33)}$$

In this final layer $L^{th}$, sigmoid activation function is utilized for non-linear transformation. From the Table 1, Sigmoid activation funtion is written as,

$$\sigma(z_i^{L+1}) = \frac{1}{1+\exp^{z_i^{L+1}}} \qquad (Eq.\ 3.34)$$

The derivative of the sigmoid function is expressed as,

$$\frac{\partial \sigma(z_i^{L+1})}{\partial(z_i^{L+1})} = \frac{\partial \frac{1}{1+\exp^{z_i^{L+1}}}}{\partial(z_i^{L+1})} \qquad (Eq.\ 3.35)$$

$$= \sigma(z_i^{L+1})(1-\sigma(z_i^{L+1})) \qquad (Eq.\ 3.36)$$

Substuting the Eqn.3.66 in Eqn.3.33,

$$\frac{\partial \mathrm{L}(\hat{y}_i^{L+1}, y_i)}{\partial w_{i,i-1}^L} = \frac{1}{t}\sum_1^t \left(\frac{-y_i}{\hat{y}_i^{L+1}} + \frac{1-y_i}{1-\hat{y}_i^{L+1}}\right)\left(\sigma\left(\sum_{i=1}^i w_{i,i-1}a^{L-1}+b^L\right)\left(1-\sigma\left(\sum_{i=1}^i w_{i,i-1}a^{L-1}+b^L\right)\right)\right) \qquad (Eq.\ 3.37)$$

where
$$\hat{y}_i^{L+1} = a_i^{L+1} = \sigma(z_i^{L+1})$$

$$\frac{\partial \mathrm{L}(\hat{y}_i^{L+1}, y_i)}{\partial w_{i,i-1}^L} = \frac{1}{t}\sum_1^t \left(\frac{-y_i}{y_i^{L+1}} + \frac{1-y_i}{1-\sigma(\sum_{i=1}^i w_{i,i-1}a^{L-1}+b^L)}\right)\left(\sigma\left(\sum_{i=1}^i w_{i,i-1}a^{L-1}+b^L\right)\left(1-\sigma\left(\sum_{i=1}^i w_{i,i-1}a^{L-1}+b^L\right)\right)\right) \qquad (Eq.\ 3.38)$$

$$\frac{\partial \mathrm{L}(\hat{y}_i^{L+1}, y_i)}{\partial w_{i,i-1}^L} = \frac{1}{t}\sum_1^t \hat{y}_i^{L+1}\left(\sigma\left(\sum_{i=1}^i w_{i,i-1}a^{L-1}+b^L - y_i\right)\right) \qquad (Eq.\ 3.39)$$

Hence, the partial derivative loss function with respect to weights of every neuron in $L^{th}$ layers is expressed as,

$$\frac{\partial \mathrm{L}(\hat{y}^{L+1}, y)}{\partial W^L} = \begin{bmatrix} \frac{\partial \mathrm{L}(\hat{y}_1^{L+1}, y_1)}{\partial w_{1,0}^L} \\ \frac{\partial \mathrm{L}(\hat{y}_2^{L+1}, y_2)}{\partial w_{2,i-2}^L} \\ \vdots \\ \frac{\partial \mathrm{L}(\hat{y}_i^{L+1}, y_i)}{\partial w_{i,i-1}^L} \\ \vdots \end{bmatrix} = \begin{bmatrix} \frac{1}{t}\sum_1^t \hat{y}_1^{L+1}(\sigma(z_1^{L+1} - y_1)) \\ \frac{1}{t}\sum_1^t \hat{y}_2^{L+1}(\sigma(z_2^{L+1} - y_2)) \\ \vdots \\ \frac{1}{t}\sum_1^t \hat{y}_i^{L+1}(\sigma(z_i^{L+1} - y_i)) \\ \vdots \end{bmatrix} \qquad (Eq.\ 3.40)$$

Partial derivative of cost function with respect to bias $b_i^l$ in $i^{th}$ neuron at $L^{th}$ layer is,

$$\frac{\partial \mathrm{L}(\hat{y}_i^{L+1}, y_i)}{\partial b_i^L} = \frac{1}{t}\sum_1^t \frac{\partial \mathrm{L}(\hat{y}_i^{L+1}, y_i)}{\partial \hat{y}_i^{L+1}} \frac{\partial \hat{y}_i^{L+1}}{\partial b_i^L} \qquad (Eq.\ 3.41)$$

$$= \frac{1}{t} \sum_1^t \frac{-y_i}{\hat{y}_i^{L+1}} + \frac{1-\hat{y}_i}{1-y_i^{L+1}}(\frac{\partial \hat{y}_i^{L+1}}{\partial b_i^L}) \qquad \text{(Eq. 3.42)}$$

$$\frac{\partial L(\hat{y}_i^{L+1}, y_i)}{\partial b_i^L} = \sigma(z_i^{L+1}) - y_i \qquad \text{(Eq. 3.43)}$$

Partial derivative of cost function with respect to bias of every neurons at $L^{th}$ is written as,

$$b^L = \begin{bmatrix} \frac{\partial L(\hat{y}_1^{L+1}, y_1)}{\partial b_1^L} \\ \frac{\partial L(\hat{y}_2^{L+1}, y_2)}{\partial b_2^L} \\ \vdots \\ \frac{\partial L(\hat{y}_i^{L+1}, y_i)}{\partial b_i^L} \\ \vdots \end{bmatrix} = \begin{bmatrix} \sigma(z_1^{L+1}) - y_1 \\ \sigma(z_2^{L+1}) - y_2 \\ \vdots \\ \sigma(z_i^{L+1}) - y_i \\ \vdots \end{bmatrix} \qquad \text{(Eq. 3.44)}$$

In this same way, partial derivatives of loss function with respect to all of hidden neruons and hidden layers can be calculated. ReLU non-linear activation function is used in all of the hidden layers from $l-1$ to $L_1$. Partial derivative of loss function with respect to weight of $i^{th}$ neuron at first layer $l$ of fully connected dense layer

$$\frac{\partial L(\hat{y}_i^{L+1}, y_t)}{\partial w_{i,i-1}^l} = \frac{\partial L(\hat{y}_i^{L+1}, y_i)}{\partial y_i^{L+1}} \frac{\partial \hat{y}_i^{L+1}}{\partial w_{i,i-1}^l} \qquad \text{(Eq. 3.45)}$$

$$= \frac{1}{t} \sum_1^t \frac{-y_i}{\hat{y}_i^{L+1}} + \frac{1-y_i}{1-\hat{y}_i^{L+1}}(\frac{\partial \hat{y}_i^{L+1}}{\partial w_{i,i-1}^l}) \qquad \text{(Eq. 3.46)}$$

$$= \frac{1}{t} \sum_1^t \frac{-y_i}{\hat{y}_i^{L+1}} + \frac{1-y_i}{1-\hat{y}_i^{L+1}}(\frac{\partial a_i^{L+1}}{\partial w_{i,i-1}^l}) \qquad \text{(Eq. 3.47)}$$

$$= \frac{1}{t} \sum_1^t \frac{-y_i}{\hat{y}_i^{L+1}} + \frac{1-y_i}{1-\hat{y}_i^{L+1}} \frac{\partial \sigma(z_i^{L+1})}{\partial w_{i,i-1}^l} \qquad \text{(Eq. 3.48)}$$

$$= \frac{1}{t} \sum_1^t \frac{-y_i}{\hat{y}_i^{L+1}} + \frac{1-y_i}{1-\hat{y}_i^{L+1}} \sigma'(z_i^l) \qquad \text{(Eq. 3.49)}$$

$$\frac{\partial L(\hat{y}_i^{L+1}, y_t)}{\partial w_{i,i-1}^l} = \frac{1}{t} \sum_1^t \frac{-y_i}{\hat{y}_i^{L+1}} + \frac{1-y_i}{1-\hat{y}_i^{L+1}} \sigma'(z_i^l) \qquad \text{(Eq. 3.50)}$$

$$\frac{\partial L(\hat{y}_i^{L+1}, y_t)}{\partial w_{i,i-1}^l} = \frac{1}{t} \sum_1^t \frac{-y_i}{\hat{y}_i^{L+1}} + \frac{1-y_i}{1-\hat{y}_i^{L+1}} \sigma'(\sum_{i=1}^i w_{i,i-1} a^{l-1} + b^l) \qquad \text{(Eq. 3.51)}$$

Since, ReLU activation function is used, than the derivative of ReLU activation function is, From the Table.1,

$$\sigma'(z) = \begin{cases} 0 & \text{if } x < 0 \\ 1 & \text{if } x \geq 0. \end{cases} \quad \text{(Eq. 3.52)}$$

If $z > 0$,

$$\frac{\partial L(\hat{y}_i^{L+1}, y_i)}{\partial w_{i,i-1}^l} = \frac{y_i - z_i^l}{z_i^l(1 - z_i^l)} \quad \text{(Eq. 3.53)}$$

Hence, partial derivative of loss function with respect to weight of all neuron at $l^{th}$ layer is,

$$W^l = \begin{bmatrix} \frac{\partial L(\hat{y}_1^{L+1}, y_1)}{\partial w_{1,0}^l} \\ \frac{\partial L(\hat{y}_2^{L+1}, y_2)}{\partial w_{2,1}^l} \\ \vdots \\ \frac{\partial L(\hat{y}_i^{L+1}, y_i)}{\partial w_{i,i-1}^l} \\ \vdots \end{bmatrix} = \begin{bmatrix} \frac{y_1 - z_1^l}{z_1^l(1 - z_1^l)} \\ \frac{y_2 - z_2^l}{z_2^l(1 - z_2^l)} \\ \vdots \\ \frac{y_i - z_i^l}{z_i^l(1 - z_i^l)} \\ \vdots \end{bmatrix} \quad \text{(Eq. 3.54)}$$

Partial derivative of loss function with respect to bias of $i^{th}$ neuron at $l^{th}$ layer is,

$$\frac{\partial L(\hat{y}_i^{L+1}, y_i)}{\partial b_i^l} = \frac{\partial L(\hat{y}_i^{L+1}, y_i)}{\partial \hat{y}_i^{L+1}} \frac{\partial \hat{y}_i^{L+1}}{\partial b_i^L} \quad \text{(Eq. 3.55)}$$

$$= \frac{1}{t} \sum_i^t \frac{-y_i}{\hat{y}_i^L + 1} + \frac{1 - y_i}{1 - \hat{y}_i^{L+1}} \left( \frac{\partial \hat{y}_i^{L+1}}{\partial b_i^l} \right) \quad \text{(Eq. 3.56)}$$

$$\frac{\partial L(\hat{y}_i^{L+1}, y_i)}{\partial b_i^l} = \sigma(z_i^{l-1}) - y_i \quad \text{(Eq. 3.57)}$$

where $\sigma$ is a ReLU non-linear activation function, hence, if $z_i > 0$,

$$\frac{\partial L(\hat{y}_i^{L+1}, y_i)}{\partial b_i^l} = z_i^{l-1} - y_i \quad \text{(Eq. 3.58)}$$

Hence, the partial derivatives of loss function with respect to bias at the layer $l$ is,

$$b^l = \begin{bmatrix} \frac{\partial L(\hat{y}_1^{L+1}, y_i)}{\partial b_i^l} \\ \frac{\partial L(\hat{y}_2^{L+1}, y_i)}{\partial b_i^l} \\ \vdots \\ \frac{\partial L(\hat{y}_i^{L+1}, y_i)}{\partial b_i^l} \\ \vdots \end{bmatrix} = \begin{bmatrix} z_1^{l-1} - y_i \\ z_2^{l-1} - y_i \\ \vdots \\ z_i^{l-1} - y_i \\ \vdots \end{bmatrix} \quad \text{(Eq. 3.59)}$$

In order to perform the learning of ConvNets, it is also neccessary to update the kernel bank weights and bias value in convolution layers as well as in pooling layers, Partial derivative of loss function with respect to input value $a_i^{l-1}$ is,

$$\frac{\partial \mathrm{L}(y_t^{L+1}, y_t)}{\partial a_i^{l-1}} = \frac{\partial \mathrm{L}(y_t^{L+1}, y_t)}{\partial y_t^{L+1}} \frac{\partial y_t^{L+1}}{\partial a_i^{l-1}} \tag{Eq. 3.60}$$

from the (Eq.1.31),

$$\frac{\partial \mathrm{L}(\hat{y}_i^{L+1}, y_t)}{\partial a_i^{l-1}} = \frac{1}{t} \sum_i^t \left(\frac{-y_i}{\hat{y}_i^L} + \frac{1-y_i}{1-\hat{y}_i^{L+1}}\right)\frac{\partial y_i^{L+1}}{\partial a_i^{l-1}} \tag{Eq. 3.61}$$

$$\frac{\partial \mathrm{L}(\hat{y}_i^{L+1}, y_i)}{\partial a_i^{l-1}} = \frac{1}{t} \sum_i^t \left(\frac{-y_i}{\hat{y}_i^L} + \frac{1-y_i}{1-\hat{y}_i^{L+1}}\right)\frac{(\partial w_{i,i-1}^{L-1} a^{L-1} + b^{L-1})}{\partial a_i^{l-1}} \tag{Eq. 3.62}$$

$$= \frac{1}{t} \sum_i^t \left(\frac{-y_{t+1}}{y_{t+1}^L + 1} + \frac{1-y_{t+1}}{1-y_{t+1}^{L+1}}\right) w_{i,i-1}^l \tag{Eq. 3.63}$$

For all input values $a^{l-1}$ at $l-1^{th}$ layer,

$$\frac{\partial \mathrm{L}(\hat{y}_i^{L+1}, y_i)}{\partial a^{l-1}} = \frac{1}{t} \sum_i^t \left(\frac{-y_{t+1}}{y_{t+1}^L + 1} + \frac{1-y_{t+1}}{1-y_{t+1}^{L+1}}\right) W^l \tag{Eq. 3.64}$$

Reshaping the long vector $\frac{\partial \mathrm{L}(y_t^{L+1}, y_t)}{\partial a^{l-1}}$

$$P^{p,q} = f^{-1} \frac{\partial \mathrm{L}(y_t^{L+1}, y_t)}{\partial a^{l-1}} \tag{Eq. 3.65}$$

Primary function of pooling layer is reduce the number of parameters and also to control the overfitting of the model. Hence, no learning takes place in pooling layers. The pooling layer error is computed by acquiring single value winning unit. Since, there are no parameters are needed to be updated in pooling layer, upsampling can be done to obtain $\frac{\partial \mathrm{L}(y_t^{L+1}, y_t)}{\partial C_{m,n}^{p,q}}$.

$$\frac{\partial \mathrm{L}(\hat{y}_i^{L+1}, y_t)}{\partial C_{m,n}^{p,q}} = P^{p,q} \tag{Eq. 3.66}$$

Partial derivative of loss function with respect to convolution kernel $k_{u,v}^{p,q}$ is,

$$\frac{\partial \mathrm{L}(\hat{y}_i^{L+1}, y_t)}{\partial k_{u,v}^{p,q}} = \sum_{m=1}^m \sum_{n=1}^n \frac{\partial \mathrm{L}(\hat{y}_i^{L+1}, y_t)}{\partial C_{m,n}^{p,q}} \frac{\partial C_{m,n}^{p,q}}{\partial k_{u,v}^{p,q}} \tag{Eq. 3.67}$$

$$\frac{\partial \mathrm{L}(\hat{y}_i^{L+1}, y_t)}{\partial k_{u,v}^{p,q}} = \sum_{m=1}^m \sum_{n=1}^n \frac{\partial \mathrm{L}(\hat{y}_i^{L+1}, y_t)}{\partial C_{m,n}^{p,q}} \frac{\sigma(\sum_{u=1}^u \sum_{v=1}^v I_{m-u,j-v} k_{u,v}^{p,q} + b^{p,q})}{\partial k_{u,v}^{p,q}} \tag{Eq. 3.68}$$

$$\frac{\partial L(\hat{y}_i^{L+1}, y_i)}{\partial k_{u,v}^{p,q}} = \sum_{m=1}^{m} \sum_{n=1}^{n} \frac{\partial L(\hat{y}_i^{L+1}, y_t)}{\partial C_{m,n}^{p,q}} I_{m-u,j-v} \quad \text{(Eq. 3.69)}$$

Updated weight of kernel $k_{u,v}^{p,q}$ can be obtained by rotating the image to 180 deg

$$\frac{\partial L(\hat{y}_i^{L+1}, y_t)}{\partial k_{u,v}^{p,q}} = \sum_{m=1}^{m} \sum_{n=1}^{n} rot I_{m-u,n-v} \cdot \frac{\partial L(\hat{y}_i^{L+1}, y_t)}{\partial C_{m,n}^{p,q}} \quad \text{(Eq. 3.70)}$$

$$k^{p,q} = rot180^o I * \frac{\partial L(\hat{y}_i^{L+1}, y_t)}{\partial C_{m,n}^{p,q}} \quad \text{(Eq. 3.71)}$$

Partial derivative of lost function with respect to bias $b^{p,q}$ of convolution kernel is,

$$\frac{\partial L(\hat{y}_i^{L+1}, y_i)}{\partial b^{p,q}} = \sum_{m=1}^{m} \sum_{n=1}^{n} \frac{\partial L(\hat{y}_i^{L+1}, y_t)}{\partial C_{m,n}^{p,q}} \frac{\partial C_{m,n}^{p,q}}{\partial b^{p,q}} \quad \text{(Eq. 3.72)}$$

$$= \sum_{m=1}^{m} \sum_{n=1}^{n} \frac{\partial L(\hat{y}_i^{L+1}, y_i)}{\partial C_{m,n}^{p,q}} \frac{\partial \sigma(\sum_{u=1}^{u} \sum_{v=1}^{v} I_{m-u,j-v} k_{u.v}^{p,q} + b^{p,q})}{\partial b^{p,q}} \quad \text{(Eq. 3.73)}$$

$$b^{p,q} = \frac{\partial L(\hat{y}_i^{L+1}, y_i)}{\partial b^{p,q}} = \sum_{m=1}^{m} \sum_{n=1}^{n} \frac{\partial L(\hat{y}_i^{L+1}, y_i)}{\partial C_{m,n}^{p,q}} \quad \text{(Eq. 3.74)}$$

## 3.3 Parameter updates

In order to minimize the loss function, it is necessary to update the learning parameter at every iteration process on the basis of gradient descent. Though various optimization techniques are developed to increase the learning speed, this article is considered only gradient descent optimization. The weight and bias update of fully connected dense layer $L+1$ is given by,

$$W^{L+1} = W^{L+1} - \alpha \frac{\partial L(\hat{y}^{L+1}, y)}{\partial W^L} \quad \text{(Eq. 3.75)}$$

$$b^{L+1} = b^{L+1} - \alpha \frac{\partial L(\hat{y}^{L+1}, y)}{\partial b^L} \quad \text{(Eq. 3.76)}$$

The weight and bias update of fully connected dense layer $l$ is given by,

$$W^l = W^l - \alpha \frac{\partial L(\hat{y}^{L+1}, y)}{\partial W^l} \quad \text{(Eq. 3.77)}$$

$$b^l = b^l - \alpha \frac{\partial L(\hat{y}^{L+1}, y_1)}{\partial b^l} \quad \text{(Eq. 3.78)}$$

The weight and bias update of convolution kernel $l$ is given by,

$$k^{p,q} = k^{p,q} - \alpha \frac{\partial L(\hat{y}^{L+1}, y)}{\partial k_{u,v}^{p,q}} \quad \text{(Eq. 3.79)}$$

$$b^{p,q} = \alpha \frac{\partial L(\hat{y}^{L+1}, y)}{\partial b^{p,q}} \quad \text{(Eq. 3.80)}$$

Where $\alpha$ is the learning rate.

# 4  Conclusion

In this article, an overview of a Convolution Neural Network architecture is explained including various activation fucntions and loss functions. Step by step procedure of feed forward and backward propogation is explained elobrately. For mathamatical simplicity concern, Grey scale image is taken as input information, kernel stride value is taken as 1, Zeropadding value is taken as 0, non-linear transformation of intermediate layer and final layers are carried out by ReLU and sigmoid activation functions. Cross entrohpy loss function is used as a performance measure of the model. However, there are numerous optimazation and regularization procedure to minimize the loss function, to increase the learning rate and to aviod the overfitting of the model, this article is an attempt of only considering the formulation of typical Convolution Neural Network architecture with graident descent optimization.